\documentclass[lettersize,journal]{IEEEtran}
\usepackage{amsmath,amsfonts}
\usepackage{algorithmic}
\usepackage{algorithm}
\usepackage{array}
\usepackage[caption=false,font=normalsize,labelfont=sf,textfont=sf]{subfig}
\usepackage{textcomp}
\usepackage{stfloats}
\usepackage{url}
\usepackage{verbatim}
\usepackage{graphicx}
\usepackage{cite}
\usepackage{color}
\usepackage[switch]{lineno}
\usepackage{multirow}
\usepackage{amssymb}
\usepackage{bbding}
\usepackage{booktabs} 
\usepackage{threeparttable}
\begin{document}

\title{Visible-Thermal Multiple Object Tracking: Large-scale Video Dataset and Progressive Fusion Approach}

\author{Yabin Zhu, Qianwu Wang, Chenglong Li, Jin Tang, Zhixiang Huang, \emph{Senior Member, IEEE}
\thanks{
Y. Zhu is with  the Key Laboratory of Intelligent Computing and Signal Processing, Ministry of Education, Anhui University, School of Public Safety and Emergency Management, Anhui University of Science and Technology, Hefei 231131, China. (email: zhuyabin0726@foxmail.com)

Q. wang is with School of Artificial Intelligence, Anhui University, Hefei 230601, China. (e-mail: wang1597474391@foxmail.com).

C. Li is with Information Materials and Intelligent Sensing Laboratory of Anhui Province, Anhui Provincial Key Laboratory of Security Artificial Intelligence, School of Artificial Intelligence, Anhui University, Hefei 230601, China. (email: lcl1314@foxmail.com)

J. Tang is with Information Materials and Intelligent Sensing Laboratory of Anhui Province, Anhui Provincial Key Laboratory of Multimodal Cognitive Computation, School of Computer Science and Technology, Anhui University, Hefei 230601, China. (email: tangjin@ahu.edu.cn)

Z. Huang is with the Anhui University, (the Information Materials and Intelligent Sensing Laboratory of Anhui Province), Center for Big Data and Population Health of IHM. (email: zxhuang@ahu.edu.cn)
}}

\markboth{Journal of \LaTeX\ Class Files,~Vol.~xx, No.~xx, July~2024}%
{Shell \MakeLowercase{\textit{et al.}}: A Sample Article Using IEEEtran.cls for IEEE Journals}


\maketitle

\begin{abstract}
The complementary benefits from visible and thermal infrared data are widely utilized in various computer vision task, such as visual tracking, semantic segmentation and object detection, but rarely explored in Multiple Object Tracking (MOT). In this work, we contribute a large-scale Visible-Thermal video benchmark for MOT, called VT-MOT. VT-MOT has the following main advantages. 1) The data is large scale and high diversity. VT-MOT includes 582 video sequence pairs, 401k frame pairs from surveillance, drone, and handheld platforms. 2) The cross-modal alignment is highly accurate. We invite several professionals to perform both spatial and temporal alignment frame by frame. 3) The annotation is dense and high-quality. VT-MOT has 3.99 million annotation boxes annotated and double-checked by professionals, including heavy occlusion and object re-acquisition (object disappear and reappear) challenges.
To provide a strong baseline, we design a simple yet effective tracking framework, which effectively fuses temporal information and complementary information of two modalities in a progressive manner, for robust visible-thermal MOT. 
A comprehensive experiment are conducted on VT-MOT and the results prove the superiority and effectiveness of the proposed method compared with state-of-the-art methods. 
From the evaluation results and analysis, we specify several potential future directions for visible-thermal MOT. 
The project is released in https://github.com/wqw123wqw/PFTrack.
\end{abstract}


\section{Introduction}
\IEEEPARstart{M}{ultiple} Object Tracking (MOT) has increasingly attracted much attention in the computer vision community due to its engineering practicality in real-world scenarios. In recent years, significant progress has been made in MOT~\cite{zhang2021fairmot, sun2020transtrack, meinhardt2022trackformer, zeng2022motr, zhang2023motrv2}. However, it still faces formidable challenges, particularly in complex environments such as low illumination, smog and haze.
To address these challenges, integrating visible and thermal infrared data has emerged as a promising solution. Visible images provide rich color and texture information but suffer from poor data quality in low illumination and haze. In contrast, thermal infrared data exhibit good quality in such environments but lack color and texture information. The robustness of MOT in complex environments can be enhanced by leveraging the complementary information of visible and thermal infrared data. Some samples are shown in Fig.~\ref{fig:samples}. 

\begin{figure}[t]
  \centering
\includegraphics[width=\linewidth]{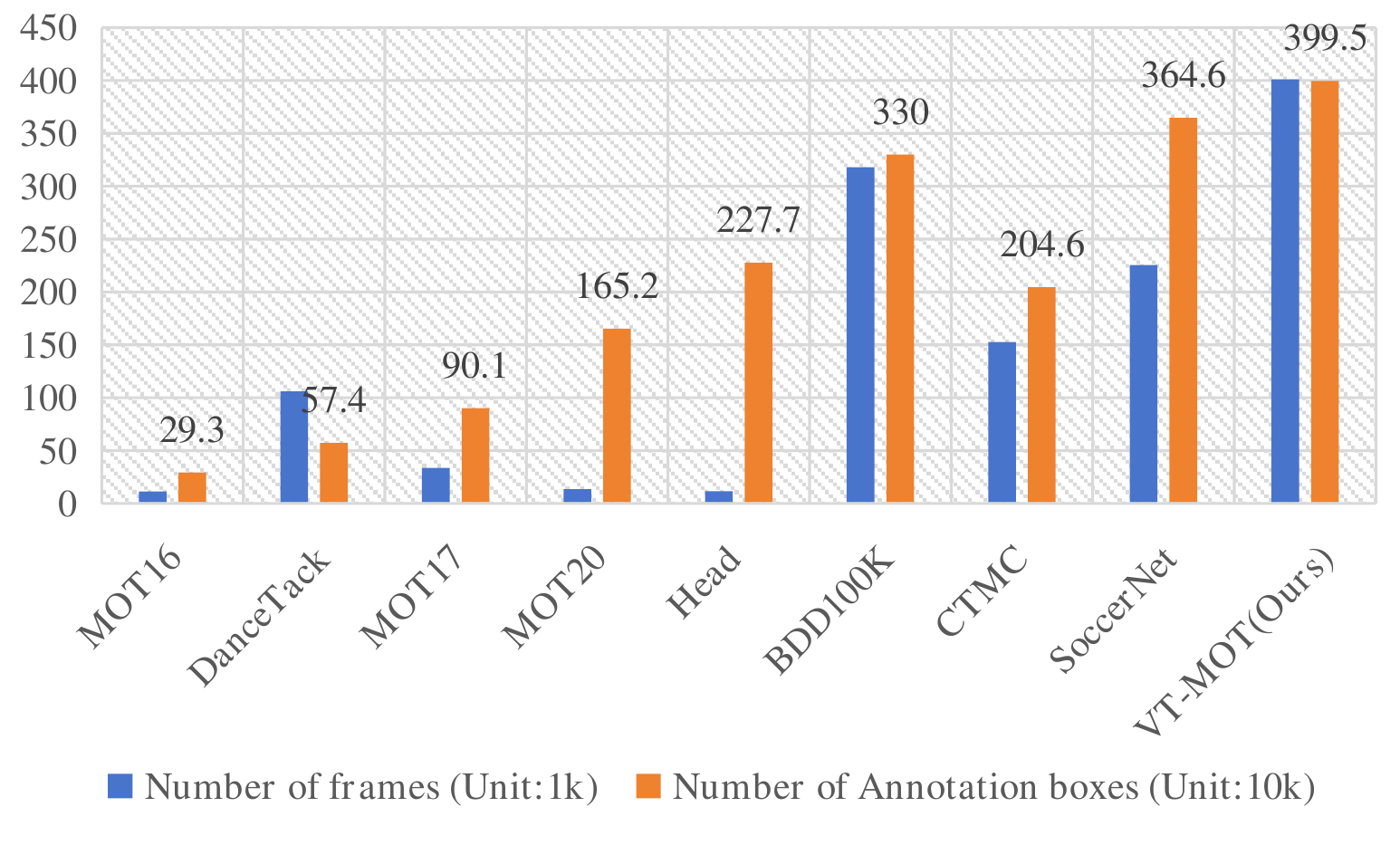}
   \caption{Comparison of our dataset with mainstream multiple object tacking datasets in terms of the number of frames and annotated bounding boxes. The data volume units for frames and annotated bounding boxes are 1k and 10k, respectively. Here, this BDD100K is the MOT subset of BDD100K.}
   \label{fig:number}
\end{figure}

Numerous studies have amply demonstrated that integrating visible and thermal infrared data can significantly improve the performance of single object tracking ~\cite{zhu2021rgbt,tang2023exploring,zhu2020quality},  semantic segmentation~\cite{frigo2022doodlenet,zhao2023mitigating,li2022rgb}, saliency detection~\cite{10003255,tu2022weakly} and object detection~\cite{zhang2023illumination,zhang2023drone} algorithms. In these research domains, several datasets have been created to promote investigations. For instance, benchmark datasets GTOT~\cite{li2016learning}, RGBT210~\cite{li2017weighted}, RGBT234~\cite{li2019rgb}, LasHeR~\cite{li2021lasher} and VTUAV~\cite{zhang2022visible} are designed specifically for single object tracking, while datasets MFNet~\cite{ha2017mfnet}, PST900~\cite{shivakumar2020pst900}, and MVSeg~\cite{ji2023multispectral} are for semantic segmentation. Furthermore, datasets VT5000~\cite{9767629}, VT1000~\cite{tu2019rgb} and VT821~\cite{wang2018rgb} are used for saliency detection. Benchmarks LLVIP~\cite{jia2021llvip}, KAIST~\cite{hwang2015multispectral}, Multispectral~\cite{takumi2017multispectral} and M3FD~\cite{liu2022target}, on the other hand, are focused on object detection.
These benchmark datasets have greatly facilitated research efforts in their respective fields. In the field of MOT, there is extremely little work utilizing visible and thermal infrared data. 
In recent years, Gebhardt et al.\cite{gebhardt2018camel} propose a dataset that can be used for visible-thermal multiple object detection and tracking. However, it has a limited scale, only containing 26 video sequences, totaling approximately 44k frames, and only about 130k annotation boxes. In addition, the lack of training sets and sufficient experimental comparisons limits the research and development of this research field. 

To handle this problem, we build a large-scale visible-thermal video benchmark for MOT. This built dataset has following major properties.
\begin{itemize}
\item{\bf Large-scale and high diversity.} Our dataset has 582 video sequence pairs with 401k frame pairs, and is collected from three platforms, including unmanned aerial vehicle (UAV), surveillance camera, and handheld device.

\item{\bf Spatio-temporal cross-modal alignment.} As multi-sensor devices cannot ensure data alignment between two modalities, we perform temporal and spatial alignment on all video sequences frame-by-frame. 

\item{\bf High-quality dense annotation.} During the annotation process, we make every effort to label every object in each frame as accurate as possible. As shown in Fig.~\ref{fig:number}, our dataset contains 3.99 million high-precision annotated boxes. 
To our best knowledge, our dataset has the highest number of annotations among MOT datasets in natural scenes.

\end{itemize}

To provide a strong baseline for visible-thermal MOT, we propose a novel progressive fusion tracking framework called PFTrack, which effectively fuses temporal information and complementary information of two modalities in a progressive manner. Previous MOT methods~\cite{zhou2020tracking, meinhardt2022trackformer, zhang2023motrv2, 10571840,
9984177} do not explore fusion techniques of visible and thermal infrared data for MOT, while single object tracking methods based on visible and thermal infrared fusion~\cite{zhu2020quality, zhu2021rgbt, liu2023quality,10506555} usually fail to the analysis and utilization of temporal information to associate multiple objects.
To fully exploit both multi-modal and temporal information for robust MOT, we design a progressive fusion module (PFM), which can be divided into two fusion stage, including temporal feature fusion and multimodal feature fusion.

For object tracking task, mining temporal information is crucial for performance improvement. Zhou et al.~\cite{zhou2020tracking} only use additive operations to fuse temporal features. However, since the current frame is not strictly spatially aligned with the previous frame, it is difficult to effectively integrate the information from previous frame.
To overcome this problem, in the first fusion stage of PFM, we employ an attention mechanism that does not rely on the strict spatial alignment of adjacent frames to integrate the temporal information well. In specific, we design the cross-attention module to fuse the features of the current and previous frames to capture the spatio-temporal contextual information. To enhance the target's localization ability, the heatmap of the previous frame is taken as the positional condition and integrated into the fused features and eventually passed through a feed-forward network to obtain the final temporal features to enhance the target feature representation. 

In multimodal feature fusion, effective intermodal interactions are crucial for enhancing the feature representation. Although the cross-attention module is a common approach to directly interact two modalities, it tends to enhance inter-modal similarity information (i.e, homogeneous information) and may ignore modality-specific information (i.e, heterogeneous information). Therefore, in the second fusion stage of PFM, we use an additive operation to obtain a rough multimodal feature, which is used as a bridging feature to interact with the unimodal feature, avoiding the problems caused by direct interaction of unimodal features. Specifically, we use four cross-attention modules to interact the fused feature and the unimodal feature as a way to further enhance the fused feature and the modality-specific feature. Finally, the enhanced features obtained from these four interactions are concatenated and fed into the feed-forward neural network to obtain a powerful multimodal feature representations.

In summary, the main contributions of this work are given as follows.
\begin{itemize}
\item We build a large-scale visible-thermal MOT dataset VT-MOT, which can promote the research and development of MOT in all weather and all day. The dataset includes 582 video sequence pairs with 401k frame pairs captured in surveillance, drone and handheld platforms. 
\item  We perform manual spatio-temporal alignment of all video sequences of both modalities in a frame by frame manner to ensure high-quality alignment of the two modalities. Moreover, dense and high-quality annotation is provided for comprehensive evaluation of different MOT algorithms. These annotation contain 3.99 million bounding boxes and heavy occlusion and object re-acquisition  challenge labels.
\item We also propose a simple yet effective progressive fusion tracking framework, which effectively fuses temporal and complementary information of two modalities in a progressive manner, for robust visible-thermal MOT.
\item We perform numerous experiments on VT-MOT dataset, and the results prove the superiority and effectiveness of the
proposed method compared with state-of-the-art methods.
\end{itemize}

\section{Related work}

\subsection{Visible-Thermal Vision Dataset}
Numerous studies have shown that the integration of visible and thermal infrared data can significantly enhance the performance of various computer vision tasks, including single object tracking, semantic segmentation, saliency detection and object detection.
To facilitate research in these domains, several benchmark datasets have been created. For single object tracking, benchmark datasets such as GTOT~\cite{li2016learning}, RGBT210~\cite{li2017weighted}, RGBT234~\cite{li2019rgb}, LasHeR~\cite{li2021lasher} and VTUAV~\cite{zhang2022visible} have been specifically designed. These datasets provide ground truth annotations for evaluation purposes.
Regarding semantic segmentation, datasets like MFNet~\cite{ha2017mfnet} and PST900~\cite{shivakumar2020pst900} have been developed to enable researchers to explore the effectiveness of visible and thermal infrared fusion in this task. 
In the area of saliency detection, datasets such as VT5000~\cite{tu2022rgbt}, VT1000~\cite{tu2019rgb}, and VT821~\cite{wang2018rgb} serve as valuable resources for investigating the fusion of visible and thermal infrared cues for salient object detection.
Furthermore, in the field of object detection, benchmarks LLVIP~\cite{jia2021llvip}, KAIST~\cite{hwang2015multispectral}, Multispectral~\cite{takumi2017multispectral}, and M3FD~\cite{liu2022target} have been established. These datasets focus on evaluating the performance of object detection algorithms using visible and thermal infrared data.

However, there are still relatively few studies utilizing visible and thermal infrared data in MOT. Gebhardt et al.~\cite{gebhardt2018camel} build a dataset suitable for visible-thermal multiple object detection and tracking. Unfortunately, this dataset has a limited scale with only 26 video sequences, totaling approximately 44k frames, and around 130k annotation boxes. Moreover, limited training sets and insufficient experimental comparisons have resulted in insufficient attention to this work. Thus, there is a need for further exploration and improvement of visible-thermal MOT methods.
\subsection{Multiple Object Tracking}
Multiple object tracking is a direction that has long existed, but past research has primarily focused on single object tracking. It is only in recent years that researchers have started to closely pay attention to MOT and have made significant breakthroughs. SORT~\cite{bewley2016simple} follows the tracking-by-detection strategy, which first utilizes a detector to detect targets and then employs Kalman filter~\cite{welch1995introduction} and Hungarian algorithm~\cite{kuhn1955hungarian} for tracking. Building upon the SORT method, DeepSORT~\cite{wojke2017simple} integrates more accurate metric combining appearance and motion information for enhancing robustness against missing data and occlusions. JDE~\cite{wang2020towards}, FairMOT~\cite{zhang2021fairmot} and CenterTrack~\cite{zhou2020tracking} further explore the joint learning of object detection and tracking. TransTrack~\cite{sun2020transtrack}, TrackFormer~\cite{meinhardt2022trackformer} and MOTR~\cite{zeng2022motr} propose more elegant end-to-end multiple object tracking framework based on attention mechanism. These methods use the feature of current frame as the key, and combines the object feature query from the previous frame and a set of learned object feature queries from the current frame as the input query for the entire network. This makes it possible to keep track of existing objects as well as emerging objects. In addition, based on MOTR, MOTRv2~\cite{zhang2023motrv2} utilizes the object boxes obtained from a YOLOX~\cite{ge2021yolox} detector as an additional proposal query for the tracking network, which can significant improve query reliability and enhance tracker performance. 
Despite the breakthrough progress achieved by the aforementioned methods, effectively tracking objects remains challenging in environments with extremely low illumination, illumination variations.
\begin{table}[h]\footnotesize 
\setlength{\belowcaptionskip}{0cm}
\caption{Detailed show of the data collected and shot in the VT-MOT.}
\centering
\renewcommand\arraystretch{1.5}
\begin{tabular}{c | c c c c} 
\hline
Data Sources &Videos  &Total Frames  & Tracks &Annotation boxes \\
\hline
Collection  &225 &95711  &4811 &1023562 \\
Shoot  & 357&305357  & 8620 & 2971215 \\\hline
\end{tabular}
\label{tb::sources}
\end{table}

\begin{table*}[t]\footnotesize 

\setlength{\belowcaptionskip}{0cm}

\caption{Comparison of our VT-MOT with public MOT datasets.}
\centering
\renewcommand\arraystretch{1.5}
\begin{tabular}{ c | c c c c c c c c}
\hline
Dataset  &Modality &Videos  &Total Frames &Frames rate  &Avg length(s) & Tracks &Annotation boxes &Density \\
\hline
MOT16~\cite{milan2016mot16} & Visible& 14 & 11235 & {14 30} & 33.071 & 1342 & 292733 & 25.80\\
MOT17~\cite{milan2016mot16} & Visible &42 & 33705& 14~30&33.07 &3993 &901119 &26.50\\
MOT20~\cite{dendorfer2020mot20} & Visible &8 & 13410& 25& 66.875& 3457&1652040 &121.28\\
KITTI-T~\cite{voigtlaender2019mots} & Visible &50 & 10870& -& -& 977&65213 &-\\
Head~\cite{sundararaman2021tracking} & Visible &9 &11463 & -& -&5230 &2276838 &-\\
CTMC~\cite{anjum2020ctmc} & Visible &86 &152584 & -& 59.11& 2900& 2045834&13.20\\
SoccerNet~\cite{cioppa2022soccernet} & Visible &201 & 225375&- & 30& 5009&3645661 &-\\
DanceTrack*~\cite{sun2022dancetrack} & Visible &100 &105855 &20&52.93 & 990& 574078 &-\\
BDD100K(MOT subset)~\cite{yu2020bdd100k} & Visible & 2000&318000 &5&40 &130600&3300000 &-\\
TAO~\cite{dave2020tao} & Visible &2907 & 4447038&- & 36.8&16104 & 332401&-\\
SportsMOT~\cite{cui2023sportsmot} & Visible &240 & 150379&25 & 25.06&3401 & 1629490&-\\
\hline 
CAMEL~\cite{gebhardt2018camel}&Visible-Thermal&26 &44500 & -&- & 800& 131940&-\\
VT-MOT-testing(Ours) &Visible-Thermal &120 &83027&25 &27.676 &2671 & 830250&10.000\\
VT-MOT-training(Ours) &Visible-Thermal &462 &318041 &25 & 27.536&10760 &3164527 &9.950\\
VT-MOT(Ours) &Visible-Thermal &582 &401068 &25 &27.57 &13431 & \textbf{3994777} &9.960 \\\hline
\end{tabular}
\begin{tablenotes}
    \item DanceTrack*: annotation boxes only statistics of training and validation sets.
\end{tablenotes}   
\label{tb::datasets}
\end{table*}

\section{Visible-Thermal Video Benchmark}
\begin{figure}[ht]
  \centering
\includegraphics[width=\linewidth]{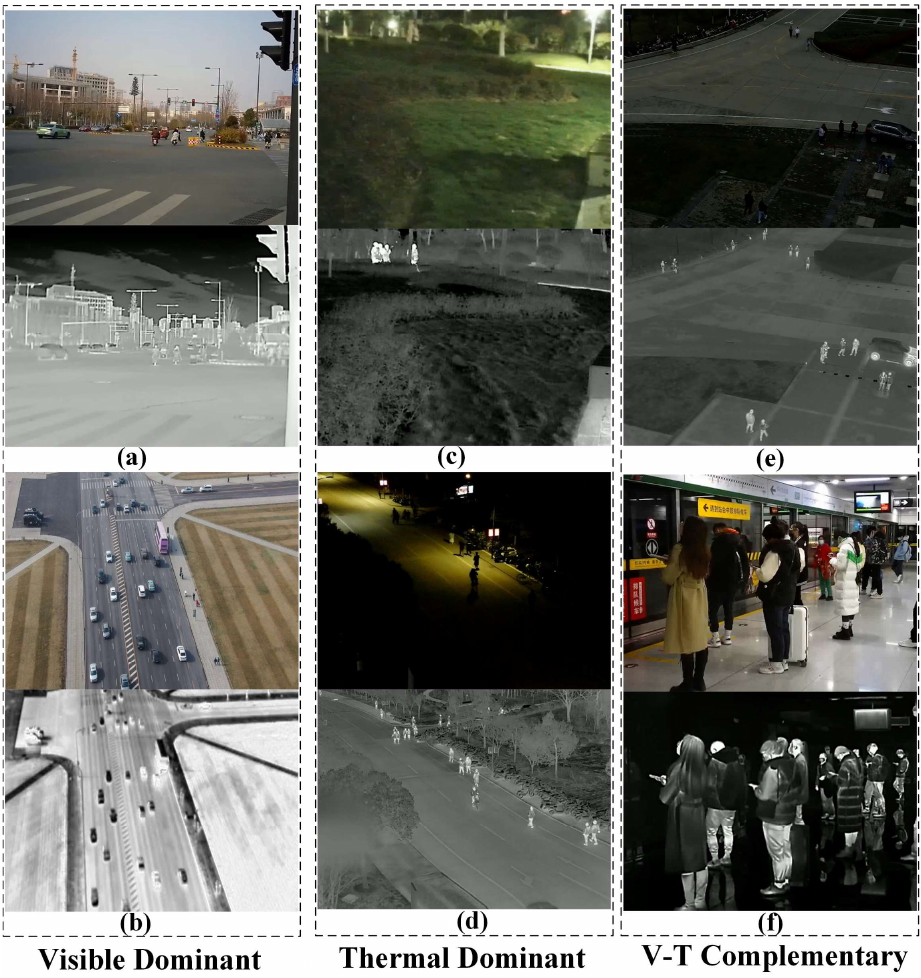}
   \caption{Some sample frames in VT-MOT.}
   \label{fig:samples}
\end{figure}

\begin{figure}[t]
  \centering
\includegraphics[width=\linewidth]{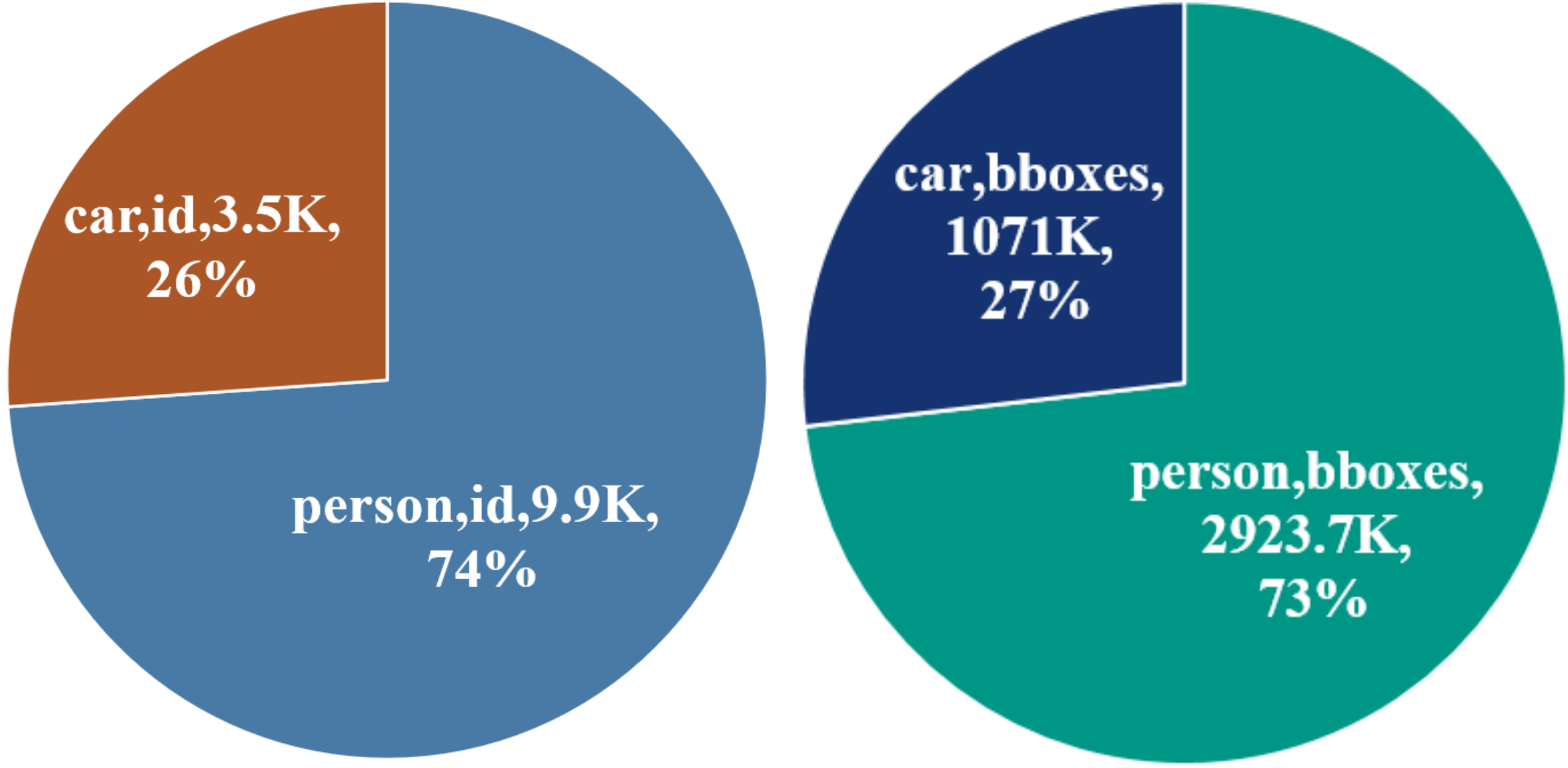}
   \caption{The number and percentage of IDs and boxes for each category in entire VT-MOT.}
   \label{fig:cls}
\end{figure}

In this section, we introduce the details of our large-scale visible-thermal video benchmark, including data collection, multi-platform imaging setup, data format, spatio-temporal cross-modal alignment, high-quality dense data annotation and data statistics.
\subsection{Data Collection}
We use HIKMICRO HM-TP2ZL-HD10 to shoot handheld scene data of two modalities.
The thermal infrared image captured by this camera has a resolution of $640\times480$, while the visible image has an even higher resolution of up to $1600\times1200$.
The DJI Matrice 300 RTK drone with a ZENMUSE H20T camera is used to capture data of the drone. The thermal infrared image has the resolution of $640\times512$, while the visible image has the resolution of $640\times512$. 
Furthermore, we use Hikvision  DS-2TD4136-25/V2 to collect data from monitoring scenes. The resolution of visible and thermal infrared modalities are $1920\times1080$ and $384\times288$, respectively.
In addition, we collect a portion of the dataset from the existing single object tracking datasets RGB-T234, LasHeR, and VTUAV. For this portion of the dataset, we re-annotate it to support multiple object tracking. 

In Table~\ref{tb::sources}, we provide detailed information about these data. 
It is evident that the shoot data has a much larger number of frames and annotated bounding boxes compared to the collected data.
As shown in Table~\ref{tb::datasets}, the VT-MOT has 582 video sequence pairs with 401k frame pairs and 3.99 million annotation boxes. To our best knowledge, our annotation boxes are the most abundant in real-world multiple object tracking datasets. The dense annotation boxes are crucial for both training deep networks and accurately evaluating their performance. To train the visible-thermal trackers, we divide the dataset into a training set and a testing set at a 4:1 ratio of the platform. The training set consists of 462 sequences, while the testing set contains 120 sequences.

In addition, in the Fig.~\ref{fig:samples}, we show some sample frames to provide viewers with a preview of our dataset. Notably, in this figure, we show three main types of samples, \emph{i.e.}, visible dominant, thermal dominant and visible-thermal complementary. Specifically, in Fig.~\ref{fig:samples} (a) and Fig.~\ref{fig:samples} (b), thermal modality has low contrast, similar object and lack of color information, on the contrary the visible modality is of good quality. In Fig.~\ref{fig:samples} (c) and Fig.~\ref{fig:samples} (d), visible modality suffers from strong illumination and extreme low illumination challenge respectively, while thermal infrared modality is undisturbed. In Fig.~\ref{fig:samples} (e), visible  modality suffers from low illumination challenge, but still can observe some object with obvious color difference, in contrast, thermal modality can capture all objects well, but lack of color information and high object similarity, therefore the two modalities have a complementary relationship. Similarly, in Fig.~\ref{fig:samples} (f), the visible modality objects are rich in color and texture information but have noisy backgrounds, in contrast, the thermal modality has prominent foreground information but lacks color information, so the two modalities also have good complementary relationship.

\begin{table*}[h]
\setlength{\tabcolsep}{4mm}
\caption{Comparison of data from different platforms in VT-MOT dataset.}
\centering
\renewcommand\arraystretch{1.5}
\begin{tabular}{ c | c c c c c c c}
\hline
Platform   &Videos  &Total Frames &Frames rate  &Avg length(s) & Tracks &Annotation boxes &Density \\
\hline
UAV  &111 & 108305&25 &39.029 &3314 &1574091 &14.53\\
surveillance  & 203&83236 &25&16.401 &4423 &881120 &10.58 \\
handheld  & 268&209527 &25 & 31.273&5694 & 1539566 &7.35 \\\hline
\end{tabular}
\label{tb::platform}
\end{table*}

\subsection{Multi-platform Imaging Setup}
Unlike other single platform MOT datasets~\cite{milan2016mot16, dendorfer2020mot20, cioppa2022soccernet, sun2022dancetrack, cui2023sportsmot}, our dataset is collected from handheld devices, drones, and surveillance platforms. Overall, it has 111 drone video pairs, 203 surveillance video pairs, and 268 handheld devices video pairs. Due to different imaging devices, the shooting time and target density of each platform vary. The dataset includes data from various perspectives and environmental conditions, which can help broaden the application scenarios of visible-thermal tracking. Handheld devices provide usage scenarios that are closer to the real world, capturing activities and dynamic changes in people’s daily lives. Drones offer an aerial perspective and a wide field of view, facilitating the tracking of targets’ movements and behaviors over a large area. Surveillance platforms provide data from specific environments, such as surveillance cameras in public spaces. By combining data from these different sources, we can significantly enhance the robustness and generalization capabilities of trackers, enabling them to adapt to various complex application scenarios. 
The details of the dataset from each platform of our dataset are presented in Table~\ref{tb::platform}.
\subsection{Data Format} 
Referring to the file format and data structure of MOT20, we designe our own MOT data format. Maintaining the consistency of unimodal and multimodal data formats will be more convenient when evaluating the tracker performance.
In detail, the VT-MOT data format is organised as follows: all video sequences are stored in a dedicated folder and named after the corresponding video. In order to distinguish between image sequences of different modalities, two subfolders, "visible" and "infrared", are created in each sequence folder. For each modal video sequence, the images are renamed by frame ID and stored in JPEG format. For example, if a 30-second video sequence contains images named 000001.jpg to 000750.jpg, then other modal images with the same timestamps will have the same filenames except for the folder names.
As for the ground truth, they are placed in subfolders within each sequence folder and documented using a comma delimited txt file containing 9 columns. These columns identify the frame ID, the track ID, the x-coordinate and y-coordinate of the upper left corner of the bounding box, the width, the height, the box validity (i.e. 1 or 0), the category labels (i.e. 1 or 2), and a fixed number 1. Additionally, a configuration file, seqinfo.ini, is provided, which contains important information such as the name, the frame rate, the resolution, the folder path and the duration of each sequence, so that the user to quickly understand the details of the dataset.

\begin{figure}[ht]
  \centering
\includegraphics[width=\linewidth]{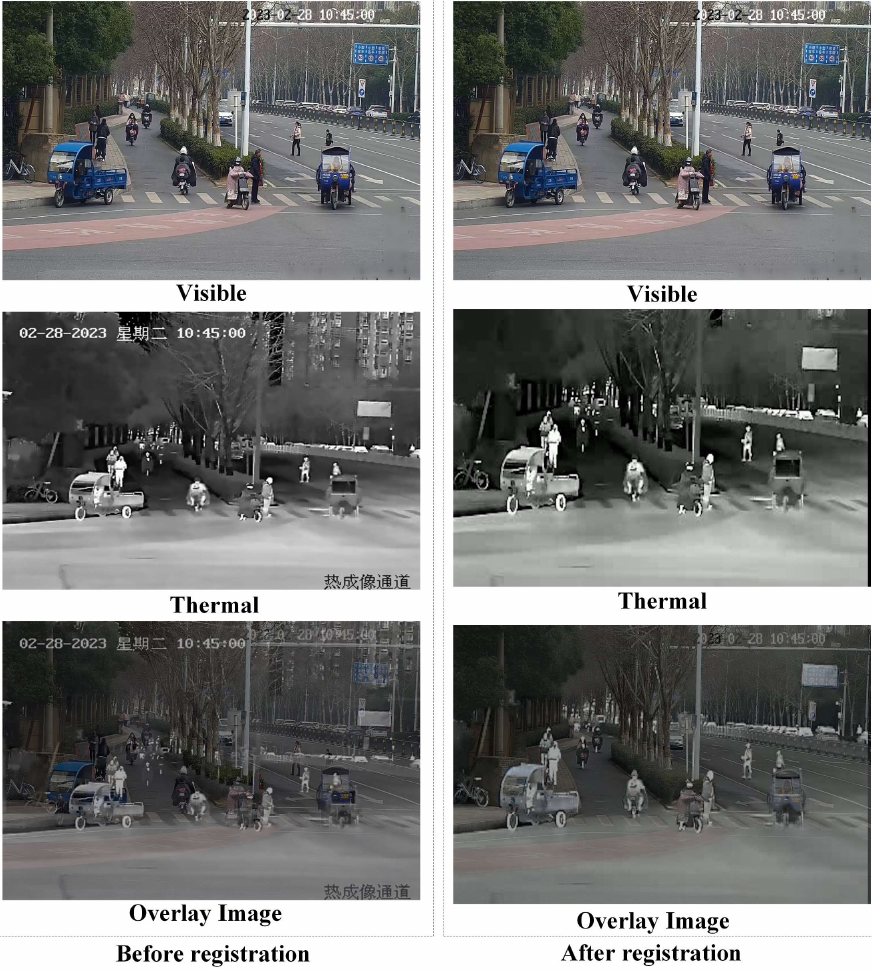}
   \caption{Registration samples.}
   \label{fig:registration}
\end{figure}

\subsection{Spatio-temporal Cross-modal Alignment}
In this dataset, there is an alignment difference between the visible light and thermal infrared modalities obtained from real-world video recordings. 
To address this issue, software Honeyview and Adobe Premiere Pro 2021 are used to manually align video clips in both modalities.
Aligning visible and thermal infrared data holds two significance. Firstly, it contributes to the ongoing research on modal fusion MOT. By manually aligning the data, we can effectively leverage the advantages of visible and thermal modalities, enhancing the accuracy of object detection and tracking.
Secondly, the manually aligned data serves as ground truth supervision for modeling modality alignment in non-aligned visible and thermal infrared MOT, facilitating subsequent research. 
To allow the reader to visualize the manual alignment results, we show an example in Fig. \ref{fig:registration}.
\begin{table*}[h]
  \centering
  \renewcommand\arraystretch{1.6}
  \caption{Comparison of data for different object scales in VT-MOT dataset.}
  \setlength{\tabcolsep}{4mm}{
    \begin{tabular}{l|ccc|cc|r}
    \hline
    & \multicolumn{3}{c|}{Small} & \multicolumn{2}{c|}{Mid} & \multicolumn{1}{c}{Large}\\
    \hline
    Interval & \multicolumn{1}{c|}{(0,11×11]} & \multicolumn{1}{c|}{(11×11,22×22]} & (22×22,32×32] & \multicolumn{1}{c|}{(32×32,64×64]} & (64×64,96×96] & \multicolumn{1}{c}{(96×96,$\infty$)} \\
    \hline
    Number of boxes & \multicolumn{1}{c|}{160964} & \multicolumn{1}{c|}{889828} &  \multicolumn{1}{c|}{844465}     & \multicolumn{1}{c|}{1194084} &    \multicolumn{1}{c|}{394951}   &  \multicolumn{1}{c}{510485}\\
    \hline
    Ratio of boxes(\%) & \multicolumn{1}{c|}{4} & \multicolumn{1}{c|}{22} & \multicolumn{1}{c|}{21}      & \multicolumn{1}{c|}{30} &     \multicolumn{1}{c|}{10} & \multicolumn{1}{c}{13} \\
    \hline
    Number of total boxes & \multicolumn{3}{c|}{1895257} & \multicolumn{2}{c|}{1589035} & \multicolumn{1}{c}{510485} \\
    \hline
    Ratio of total boxes(\%) & \multicolumn{3}{c|}{47} & \multicolumn{2}{c|}{40} &  \multicolumn{1}{c}{13}\\
    \hline
    \end{tabular}}%
  \label{size-detail}%
\end{table*}%
\subsection{High-quality Dense Annotation} 
In order to ensure that the labeling of MOT is done with high quality, we take the following measures. We recruit forty-seven volunteers to participate in the annotation work, increasing the labor resources of the annotation team and improving efficiency. 
We also assign two dedicated annotation inspectors who are responsible for checking the quality of each frame’s annotation. 
Specifically, we use the ViTBAT~\cite{biresaw2016vitbat} annotation software and spent several months annotating, completing approximately 3.99 million bounding box annotation to the best of our ability. In addition, we specially annotate the severe occlusion and object re-acquisition challenges to promote related research. We believe that this dataset will undoubtedly provide a more solid data foundation for the field of visible-thermal MOT. 

\begin{figure}[htb]
\setlength{\abovecaptionskip}{0.cm}
\includegraphics[width=\linewidth]{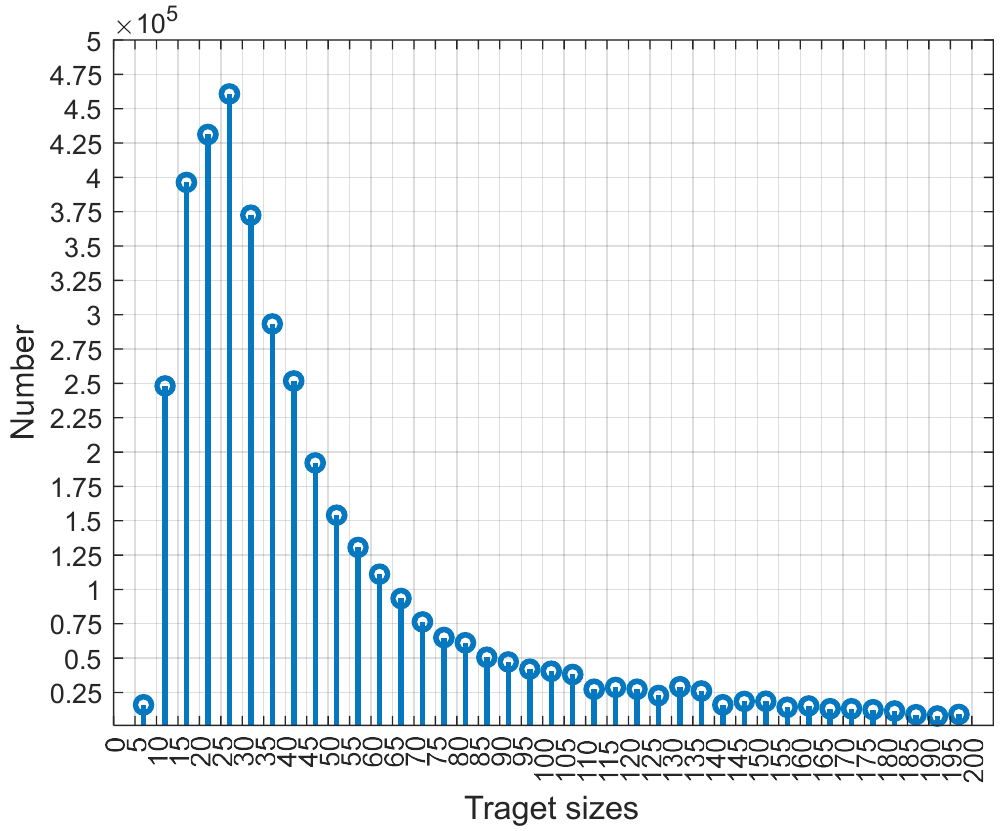}
\caption{The scale distribution of bounding boxes in our dataset. The horizontal coordinate represents the square root of the area of the bounding box. The vertical coordinate indicates the number of boxes in each scale sub-interval.}
\label{fig:distribution}
\end{figure}

\subsection{Data Statistics}
\noindent{\bf Object category.} Like most MOT datasets, our dataset prefers to focus on target categories that are common in our daily lives, i.e., people and vehicle. Such a choice is based on several factors: people and vehicle are the most common targets in our daily lives, and they are also among the most challenging targets in MOT applications. In addition, focusing on these common targets helps to better study and improve the performance of MOT algorithms for real-world applications. In Fig.~\ref{fig:cls}, we show the number and percentage of IDs and boxes for each category. Specifically, the number of vehicle IDs and boxes are 3.5k, 1070k, respectively, with a percentage of 26\% and 27\%, respectively, and the rest are the number and percentage of people IDs and boxes. 
We notice that the data distribution of VT-MOT on object categories conforms to the long-tail distribution, in which the learning under this unbalanced data is an important topic in practical applications. It can encourage the exploration of more practical and extensible tiny object tracking methods.

\noindent{\bf Scale distribution.}  
In Table~\ref{size-detail}, we also count the specific number of small, medium, and large object in our dataset based on the definition of large, medium, and small object in COCO~\cite{lin2014microsoft}. Specifically, the large object scale ranges from greater than $[96\times96, \infty)$, the medium object scale ranges from $(32\times32, 96\times96)$, and the small target scale ranges from $(0, 32\times32]$. Further, we divide the small object interval into three sub-intervals, and the medium object interval into two sub-intervals, in order to gain a more detailed understanding of the dataset scale distribution. In addition,  to give the reader an overview of the scale distribution of targets in our dataset, we also show the scale distribution map of the dataset in Fig.~\ref{fig:distribution}. From the table and figure, we can observe that the dataset has the largest number of small targets, while the number of medium and large targets decreases in order. This distributional feature can be attributed to the outdoor acquisition environment of our dataset, where the number of medium and distant videos is slightly higher. 

\begin{figure*}[t]
  \centering
\includegraphics[width=1\linewidth]{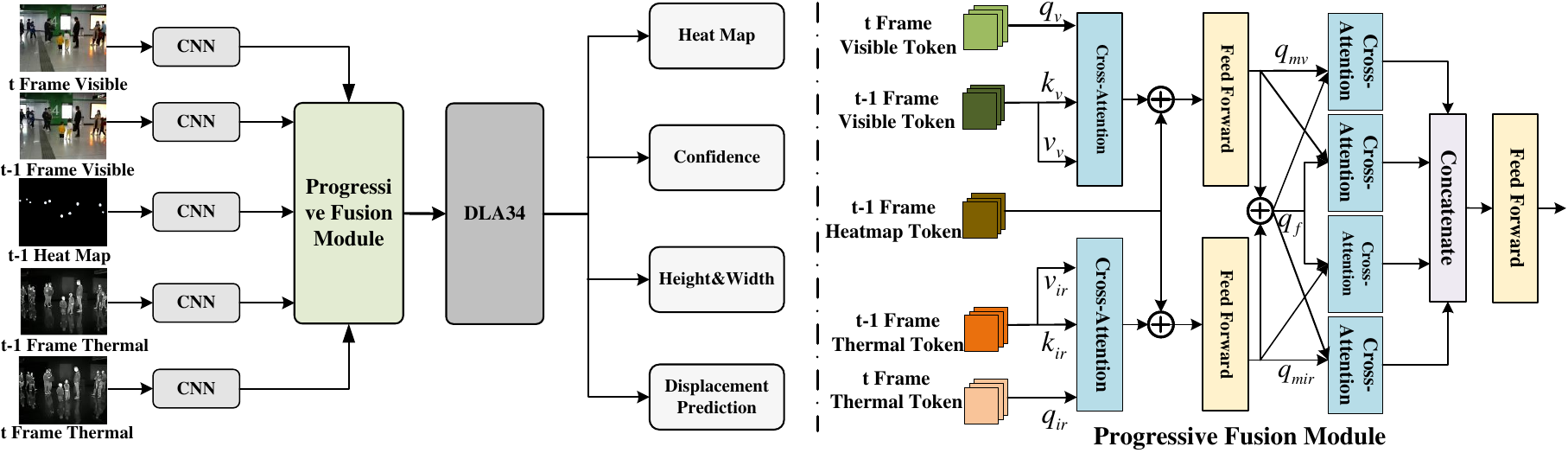}
   \caption{Our visible-thermal multiple obejct tracking framework. In order to visualize the fusion process of PFM, we omit the details of Drop, LayerNorm, residual connection and ReLU in PFM.}
   \label{fig:framework}
\end{figure*}

\section{Methodology}
In this section, we will first give an overview of our proposed method. Then, we describe the proposed progressive fusion module in detail. 
\subsection{Overview}
As shown in Fig. \ref{fig:framework}, given visible frame, thermal frame and heatmap, we first extract their features using a set of Convolutional Neural Network (CNN). In this paper, this heatmap is made from the position of the center of objects in the previous frame. This CNN consists of $7\times7$ convolutional layer, Batch Normalization (BN) layer and Rectified Linear Units (ReLU) layer. Then the obtained unimodal features containing different temporal frames are split and flattened into sequences of
patches by patch embedding layer. In this paper, this patch embedding layer is a convolutional layer with a kernel of 16 and a stride of 16. 
Next, these patches are fed into our proposed Progressive Fusion Module (PFM), which fuses the temporal and modal complementary information of two modalities to enhance the feature representation of objects. This PFM will be analyzed in detail in the next sub-section. Next, Deep Layer Aggregation Network (DLA) \cite{yu2018deep} is used as backbone network to further extract multi-modal features. Finally, four task heads are used to predict the center position and size of target. For more details, please refer to \cite{zhou2020tracking}, which is our baseline method.

\subsection{Progressive Fusion Module}
Previous MOT methods~\cite{zhou2020tracking, meinhardt2022trackformer, zhang2023motrv2} do not explore fusion techniques of visible and thermal infrared data for MOT, while single object tracking methods based on visible and thermal infrared fusion~\cite{zhu2020quality, zhu2021rgbt, liu2023quality} usually fail to the analysis and utilization of temporal information to associate multiple objects.
To fully exploit both multi-modal and temporal information for robust MOT, we propose a progressive fusion module, which is divided into two fusion stages, including temporal feature fusion and multimodal feature fusion. Mathematically, the computation of our PFM module can be written as:
\begin{equation}
\begin{aligned}
f &=\mathcal{M}(\mathcal{T}( x_{v}^{t}, x_{v}^{t-1},  x_{hm}^{t-1}), \mathcal{T}( x_{ir}^{t},  x_{ir}^{t-1},  x_{hm}^{t-1}));
\end{aligned}
\end{equation}
where $\mathcal{T}(,,)$ is the temporal feature fusion operation and the $\mathcal{M}(,)$ denotes the multimodal feature fusion operation.
For object tracking tasks, extracting temporal information is crucial for performance improvement. Our baseline method ~\cite{zhou2020tracking} only uses additive operations to integrate temporal features. However, since the current frame is not strictly spatially aligned with the previous frame, it is difficult to effectively integrate the information from the previous frame. To address this issue, in the first stage of PFM, we use an attention mechanism that does not depend on the strict spatial alignment of adjacent frames to effectively integrate temporal information well. Specifically, we first use the cross-attention module to fuse the current features (as Query (Q)) and previous frame features (as Key (k) and Value (V)) to capture the spatio-temporal contextual information. Then, to enhance the target’s localization ability, the heatmap of the previous frame is taken as
the positional condition and integrated into the fused features and eventually passed through a feed-forward network to obtain the final temporal features to enhance the target feature representation. Mathematically, the computation of our temporal feature fusion can be written as:
\begin{equation}
\begin{aligned}
 x_{v}&=CrossAttention( x_{v}^{t}+p, x_{v}^{t-1}+p, x_{v}^{t-1});\\
\bar x_{v}&= LN\left( x_{v}^{t}+ x_{v}\right)+ x_{hm}^{t-1};\\
\hat x_{v}&= LN\left(\bar x_{v}+FFN(\bar x_{v})\right);
\end{aligned}
\end{equation}
where $p$, $LN$ and $FFN$ are position encoding, LayerNorm operation and feed-forward network, respectively. $x_{v}^{t}$, $x_{v}^{t-1}$ and $x_{hm}^{t-1}$ are the features of the $t$ frame, $t-1$ frame and t-1 heatmap. Similarly, we can obtain the temporal feature $\hat x_{ir}$ of thermal infrared based on the above method.

\begin{table*}[]
\setlength{\belowcaptionskip}{2cm}
\renewcommand\arraystretch{1.6}
\caption{Compare the tracking performance of the investigated trackers on MOT17, MOT20, DanceTrack and VT-MOT.}
\setlength{\tabcolsep}{2.75mm}{
\begin{tabular}{l|ccc|ccc|ccc|ccc}
\hline

\multicolumn{1}{c|}{\multirow{2}[0]{*}{Method}}  & \multicolumn{3}{c|}{MOT17}                    & \multicolumn{3}{c|}{MOT20}                    & \multicolumn{3}{c|}{DanceTrack}   & \multicolumn{3}{c}{VT-MOT} \\ \cline{2-13} 
& HOTA          & MOTA          & IDF1          & HOTA          & MOTA          & IDF1          & HOTA          & MOTA          & IDF1          & HOTA   & MOTA   & IDF1   \\ \hline
CenterTrack~\cite{zhou2020tracking}                                  & 52.2          & 67.8          & 64.7          & -             & -             & -             & 41.8          & 86.8          & 35.7          & {39.045} &30.585 & 44.42 \\
TraDes~\cite{Wu2021TrackTD}  &    52.7      &   69.1      &  63.9        &  -      & -         &  -         &    43.3       &    86.2       &   41.2      &38.319  & 34.632& \textbf{47.008} \\
FairMOT~\cite{zhang2021fairmot}                                      & 59.3          & 73.7          & 72.3          & 54.6          & 61.8          & 67.3          & 39.7          & 82.2          & 40.8          & 37.35  & \textbf{37.266} & 45.795 \\
TransTrack~\cite{sun2020transtrack}                                   & 54.1          & 75.2          & 63.5          & 48.5          & 65.0            & 59.4          & 45.5          & 88.4          & 45.2          & 38.00 &  36.156&43.567   \\
ByteTrack~\cite{zhang2022bytetrack}                                    & 63.1          & \textbf{80.3} & 77.3          & 61.3          & \textbf{77.8} & 75.2          & 47.7          & 89.6          & {53.9} & 38.393 & 33.151 & 45.757  \\
OC-SORT~\cite{Cao2022ObservationCentricSR}    & {63.2} & 78.0      & 77.5 & {62.1} & 75.5          & {75.9} & {55.1} & \textbf{92.0} & 54.6          &   31.479     &    28.948   &    38.086    \\ 
Hybrid-SORT~\cite{yang2024hybrid}  
& \textbf{63.6} & 79.3   & \textbf{78.4}
&\textbf{62.5} & 76.4 & \textbf{76.2} 
& \textbf{62.2} & {91.6} & \textbf{63.0}  
&\textbf{39.485}     &31.074   &46.310    \\ \hline
\end{tabular}}
\label{tab:muldata-result}%
\end{table*}
It is particularly important to interact inter-modal features for obtaining an effective multimodal feature representation. Although the cross-attention module is a common approach to directly interact two modalities, it tends to enhance inter-modal similarity information ( homogeneous information) and may ignore modality-specific information ( heterogeneous information). Therefore, in the second fusion stage of PFM, we use an additive operation to obtain a rough multimodal feature, which is used as a bridging feature to interact with the unimodal feature, avoiding the problems caused by direct interaction of unimodal features. Specifically, this module initially employs a straightforward additive operation to integrate the features from two different modalities, thereby obtaining an initial multimodal features representation. Then, the fused multimodal features are used as Key (K) and Value (V), while the two unimodal features are used as Query (Q), respectively. Two cross-attention modules are used to interact the unimodal features with the multimodal features to enhance the representation of modality-specific features. At the same time, the multimodal features are also used as Q and the unimodal features as K and V. The other two cross-attention modules are then utilized to further enhance the multimodal features. Finally, the two enhanced modality-specific features and the two enhanced multimodal features obtained after these four interactions are concatenated together and then fed into a feed-forward neural network to obtain the final refined multimodal feature. Mathematically, the computation of our multimodal feature fusion can be written as:

\begin{equation}
\begin{aligned}
x_{f}&= \hat x_{v} + \hat x_{ir};\\
\bar x_{f} &=Cat(\mathcal{I}(\hat x_{v},x_{f}),\mathcal{I}(\hat x_{ir},x_{f}),\mathcal{I}(x_{f},\hat x_{v}),\mathcal{I}(x_{f},\hat x_{ir}));\\
\hat x_{f} &= LN(FFN(\bar x_{f}));
\end{aligned}
\end{equation}

where $\mathcal{I}(,)$ denotes cross-attention module.

\section{Experiment}

\begin{table*}[htbp]
\footnotesize
\centering
\renewcommand\arraystretch{1.6}
\setlength{\belowcaptionskip}{2cm}
\caption{Tracking performance comparison of several evaluated trackers on VT-MOT testing set.}
\setlength{\tabcolsep}{5.5mm}{
\begin{tabular}{l|cclllll}
\hline
\multicolumn{1}{l|}{{Method}}&Publication & {Modality} & {HOTA} & {DetA}  & {MOTP} & {IDF1} & {MOTA} \\
\hline
{FairMOT~\cite{zhang2021fairmot}} &IJCV 2021& V-T &    37.35 & 34.628       &   72.525    &  45.795   & \textcolor{blue}{\textbf{37.266}} \\
\hline
{CenterTrack~\cite{zhou2020tracking}} &ECCV 2020& V-T & \textcolor{blue}{\textbf{39.045}} &  \textcolor{blue}{\textbf{38.104}}& 72.874 & 44.42 & 30.585 \\
\hline
{TraDes~\cite{Wu2021TrackTD}} &CVPR 2021& V-T & 38.319 & 36.37   & 72.295 & \textcolor{blue}{\textbf{47.008}} &34.632  \\
\hline
{TransTrack~\cite{sun2020transtrack}} &arXiv 2021& V-T & 38.00 &  35.711  & \textcolor{blue}{\textbf{73.823}} & 43.567 &36.156  \\
\hline
{ByteTrack~\cite{zhang2022bytetrack}} &ECCV 2022& V-T &   38.393    &   32.122       &   73.483    &   45.757    &  33.151\\
\hline
{OC-SORT~\cite{Cao2022ObservationCentricSR}} &CVPR2023& V-T & 31.479 & 25.244& 73.15& 38.086&28.948  \\
\hline
{MixSort-OC~\cite{cui2023sportsmot}} &ICCV2023& V-T &39.09  &33.109 & 73.632& 45.799 &31.33  \\
\hline
{MixSort-Byte~\cite{cui2023sportsmot}} &ICCV2023& V-T &39.575  &34.806 & 73.049& 46.367 &31.593  \\
\hline
{PID-MOT~\cite{10342840}} &TCSVT 2023& V-T &35.621  &33.245 & 71.794& 42.43 &33.333  \\
\hline
{Hybrid-SORT~\cite{yang2024hybrid}} &AAAI2024& V-T &39.485  &34.619 & 72.840& 46.310 &31.074  \\
\hline
\textbf{Ours} &-& V-T &   \textcolor{red}{\textbf{41.068}}   &  \textcolor{red}{\textbf{41.631}}      &   \textcolor{red}{\textbf{73.949}}   &\textcolor{red}{\textbf{47.254}}     &  \textcolor{red}{\textbf{43.088}}\\
\hline
\end{tabular}}%
\label{tab:testing-result}%
\end{table*}%

\begin{table}[]
\footnotesize
\centering
\renewcommand\arraystretch{1.5}
\setlength{\belowcaptionskip}{-0cm}
\caption{Comparison of the performance of several trackers on unimodal data and dual-modal data sets from the VT-MOT testing set. }
\setlength{\tabcolsep}{1.6mm}{
\begin{tabular}{l|clllll}
\hline
            
\multicolumn{1}{c|}{{Method}}      & \multicolumn{1}{c}{{Modality}} & \multicolumn{1}{c}{{HOTA}} & \multicolumn{1}{c}{{DetA}} & \multicolumn{1}{c}{{IDF1}} & \multicolumn{1}{c}{{MOTP}} & \multicolumn{1}{c}{{MOTA}} \\ \hline
              & V    &            33.917    &      31.497                      &     37.966          &  71.493    &     24.266         \\
    {CenterTrack}          & V-T  &    \textbf{39.045}  &    \textbf{38.104}    &   \textbf{44.42}  &     \textbf{72.874}          &  \textbf{30.585}    \\
 \hline
              & V    &     33.308           &    28.182              &     35.962         &     72.36            &           21.575   \\
   {TransTrack}            & V-T  &  \textbf{38.00}    &    \textbf{35.711}                & \textbf{43.567}    &\textbf{73.823}     &  \textbf{36.156}    \\
  \hline
              & V   &      38.037                               &    32.003                                                                     &     44.972                                 &    \textbf{73.566}                                 &          31.808                   \\
             {ByteTrack} & V-T     &   \textbf{38.393}                         &  \textbf{32.122}                                                   &            \textbf{45.757}                 &  73.483 &                  \textbf{33.151}                    \\
 \hline
\end{tabular}}%
\label{tb:one-two-results}

\end{table}

In this section, we conduct extensive experiments on our newly proposed VT-MOT benchmark. Specifically, we will first introduce the experimental settings, including evaluation metrics, implementation details and evaluation protocols. Then, we report quantitative evaluation results, which contain comparison results with other trackers in two protocols and some ablation studies. Finally, we also give some qualitative evaluation results for visualising the performance of trackers.

\subsection{Evaluation Metrics}
To evaluate the performance of MOT algorithms on our VT-MOT dataset, we focus on two main metrics: Multi-Object Tracking Accuracy (MOTA) and High-Order Tracking Accuracy (HOTA). MOTA is a traditional benchmark for detection performance but lacks in representing association performance. To address this, Luiten et al.~\cite{Luiten2020HOTAAH} introduce HOTA, which separately evaluates detection (DetA) and association (AssA) performance, and effectively integrates both aspects into a single metric. For detailed understanding, see~\cite{Luiten2020HOTAAH}. We also use IDF1 for object association analysis, DetA for detection analysis, and MOTP for object position accuracy during tracking. We employ TrackEval\cite{trackeval} as our evaluation tool. As this evaluator can only evaluate multi-class tracking as a single class, we generate a copy of the ground truth file named gt1.txt in the test set. The only difference is that we set all class labels to 1 to facilitate evaluation.

\subsection{Implementation Details}
The experiments of the proposed tracker are conducted on 128 AMD EPYC 7542 32-Core Processor, 1 NVIDIA GeForce RTX 4090 GPUs with 24GB memory. All experiments of our method are conducted using PyTorch-1.12.1. During the training phase, we train the entire network for 10 epochs with a learning rate set to 0.000125 on the training set of VT-MOT. It is worth noting that, we adhere to the CenterTrack (our baseline) configuration without modifying any hyperparameters, except for adding a progressive fusion module and changing the number of output categories.



\subsection{Evaluation Protocols}
Considering the need for a comprehensive assessment of MOT algorithms on the VT-MOT dataset, we propose two distinct evaluation protocols.

{\bf Protocol \uppercase\expandafter{\romannumeral 1}.}
In protocol \uppercase\expandafter{\romannumeral 1}, we provide a testing set, which include 120 video sequence pairs with 83027 frame pairs from different platforms. The protocol aims to understand the overall performance and generalization capabilities of the tracker across different platforms by evaluating it on multiple platforms at the same time.

{\bf Protocol \uppercase\expandafter{\romannumeral 2}.}
In protocol \uppercase\expandafter{\romannumeral 2}, we categorize the VT-MOT testing set into three distinct groups: 58 sequences captured by handheld cameras, 40 sequences sourced from surveillance, and 22 sequences captured by drones. The Protocol \uppercase\expandafter{\romannumeral 2} is designed to evaluate the tracking metrics of each platform separately to facilitate the development of trackers on specific platforms.

\subsection{Quantitative Evaluation}
In this section, we compare our method with state-of-the-art trackers, including ByteTrack~\cite{zhang2022bytetrack}, OC-SORT~\cite{Cao2022ObservationCentricSR}, FairMOT~\cite{zhang2021fairmot}, CenterTrack~\cite{zhou2020tracking}, TraDes~\cite{Wu2021TrackTD}, TransTrack \cite{sun2020transtrack}, MixSort~\cite{cui2023sportsmot}, PID-MOT~\cite{10342840} and Hybrid-SORT~\cite{yang2024hybrid}.
To evaluate these trackers on VT-MOT, we expand them to accept two modal inputs. 
Specifically, we employ early fusion strategy to integrate the information of two modalities.
Due to the limited generalization ability of the current multiple object tracker and the significant differences between the previous dataset and VT-MOT, we have to retrain these trackers on the training set of VT-MOT to adapt them to the testing set of VT-MOT. It is important to note that we do not modify any hyper-parameters of the evaluation algorithm.
\subsubsection{Evaluation results with protocol \uppercase\expandafter{\romannumeral 1}}
In Table~\ref{tab:muldata-result}, we show the results of several tracking methods on MOT17, MOT20, DanceTrack and VT-MOT. From these results, we can observe that the performance of these trackers on the proposed VT-MOT dataset is significantly worse than their performance on other mainstream datasets. Particularly, the MOTA metric of these trackers perform worse compared to the IDF1 and HOTA metrics on the VT-MOT dataset. This may be due to several reasons. Firstly, the VT-MOT  consists of data from three platforms with different styles and more varied and complex scenarios, therefore putting a higher demand on the generalization ability of the trackers. Secondly, the VT-MOT focuses more on medium to long-range outdoor scenes, where 87\% of the targets in the video are small and medium-size targets, which poses a great challenge for detection and tracking. Lastly, the VT-MOT contains data from low-illumination scenes, which requires effective fusion of complementary information from both modalities. However, these trackers are essentially unimodal trackers and lack a well-designed fusion strategy to address the low-illumination challenge. In summary, multi-object tracking in real-world scenarios still has a long way to go. 

In Table~\ref{tab:testing-result}, we show more performance metrics of our proposed tracker and several other trackers on the VT-MOT testing set. From the results, we can observe that the performance of our method achieves the best performance in all metrics compared to the other tracking methods. In particular, it is 5.822\% higher than the best method on the MOTA metric. 


In Table~\ref{tb:one-two-results}, we compare the performance of the three trackers on unimodal and bimodal data. Note that to ensure fairness, we pre-trained these trackers on a unimodal subset of the VT-MOT when evaluating them. As can be seen from the results, CenterTrack and TransTrack perform better on bimodal data than they do on the unimodal dataset. This results can demonstrate that utilizing both visible and thermal infrared data can lead to more competitive performance in complex scenes.
\begin{table}[htbp]
  \footnotesize
\renewcommand\arraystretch{1.5}
\setlength{\belowcaptionskip}{0cm}
  \caption{Compare the performance of several trackers using early and mid-term fusion schemes in the VT-MOT testing set}
  \setlength{\tabcolsep}{0.2mm}{
    \begin{tabular}{l|p{4.19em}|lllll}
    \hline
    \multicolumn{1}{p{4.19em}|}{\textbf{Method}} & \textbf{Fusion} & \multicolumn{1}{p{4.19em}}{\textbf{HOTA}} & \multicolumn{1}{p{4.19em}}{\textbf{DetA}} & \multicolumn{1}{p{4.19em}}{\textbf{MOTP}} & \multicolumn{1}{p{4.19em}}{\textbf{IDF1}} & \multicolumn{1}{p{4.19em}}{\textbf{MOTA}} \\
    \hline
    \multicolumn{1}{l|}{\multirow{2}[0]{*}{FairMOT}} & early & 34.552 & 34.272 & 71.588 & 40.73 & 34.184 \\
         & mid   & \textbf{37.35} & \textbf{34.628} & \textbf{72.252} & \textbf{45.795} & \textbf{37.266} \\
    \hline
    \multicolumn{1}{l|}{\multirow{2}[0]{*}{CenterTrack}} & early &\textbf{39.045} & \textbf{38.104} & \textbf{72.874} & \textbf{44.42} & \textbf{30.585} \\
         & mid   &     36.895  & 35.629      &  72.291   &    41.767   &26.364  \\
   \hline
    \multicolumn{1}{l|}{\multirow{2}[0]{*}{ByteTrack}} & Early & 37.825 & 32.093 & \textbf{73.499} & 44.95 & 32.604 \\
   & Mid   & \textbf{38.393} & \textbf{32.122} & 73.483 & \textbf{45.787} & \textbf{33.151} \\
    \hline
    \multicolumn{1}{l|}{\multirow{2}[0]{*}{OC-SORT}} & Early & 29.026 & 23.13 & \textbf{73.236} & 33.486 & 23.965 \\
         & Mid   & \textbf{31.479} & \textbf{25.244} & 73.15 & \textbf{38.086} & \textbf{28.948} \\
    \hline
    \multicolumn{1}{l|}{\multirow{2}[0]{*}{Hybrid-SORT}} & Early & 39.204 & 34.516 & \textbf{72.888} &46.033 & 30.49 \\
         & Mid   & \textbf{39.485} & \textbf{34.619} & 72.84 & \textbf{46.31} & \textbf{31.074} \\
    \hline    
    {Ours} & PFM & \textcolor{red}{\textbf{41.068}} &\textcolor{red}{\textbf{41.631}} & \textcolor{red}{\textbf{73.949}} & \textcolor{red}{\textbf{47.254}} & \textcolor{red}{\textbf{43.088}} \\
    \hline
    \end{tabular}}%
  \label{tab:fusion}%
\end{table}%

In order to compare the performance of different fusion strategies in visible-thermal MOT task, we implement early and mid-term fusion strategies for several comparison algorithms. As shown in Table~\ref{tab:fusion}, for the vast majority of trackers, the mid-term fusion strategy provides a significant performance improvement compared to the early fusion strategy. However, there are exceptions, such as the CenterTrack, where the early fusion strategy outperforms the mid-term fusion strategy in specific cases. These findings help us to gain a deeper understanding of the specific impact of different fusion strategies on different MOT. In addition, compared with other fusion methods, our fusion method has a significant performance advantage. These results validate the effectiveness and excellence of our fusion method.
\subsubsection{Evaluation results with protocol \uppercase\expandafter{\romannumeral 2}}
\begin{figure*}[t]

\setlength{\abovecaptionskip}{0.cm}
\includegraphics[width=0.90\textwidth]{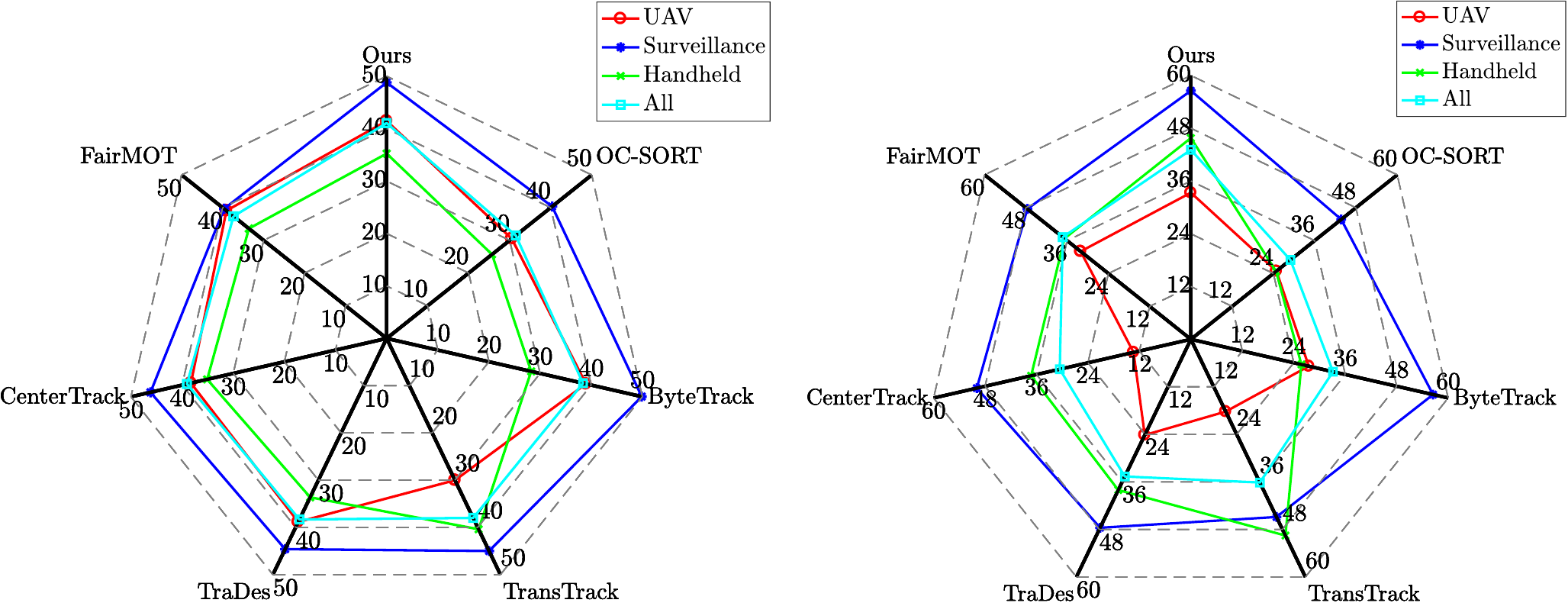}
   \caption{We compare the HOTA and MOTA scores of several trackers on UAV, surveillance and handheld platforms. The "All" denotes the integrated  platform (i.e. the entire VT-MOT testing set).}
   \label{fig:platform-hota-mota}
\end{figure*}

In order to better evaluate the performance of the trackers on different platforms, we present the HOTA and MOTA scores of our method with several trackers on UAV, handheld, and surveillance platforms in Fig.~\ref{fig:platform-hota-mota}, respectively. From these results we observe the following. 

Firstly, our method has competitive performance on UAV and surveillance platforms. The scenarios faced by these platforms usually contain numerous small objects. We design a progressive fusion moudle that belongs to the early fusion techniques, which are able to mine the contextual information of small objects more effectively. In addition, we use DLA34 as the backbone network, which is particularly well suited for extracting high-resolution features, thus ensuring high-quality feature representation of small objects. 
Secondly, it can be seen from the MOTA score that our method is more robust on a comprehensive platform.
Finally, several trackers perform best on surveillance platform data, possibly because surveillance platform can provide more stable image capture compared to drones and handheld devices. 

\begin{table}[ht]
  \centering  
  \normalsize
  \caption{Ablation studies of different components of the proposed method on VT-MOT testing set.}
  \resizebox{\linewidth}{!}{
      \renewcommand\arraystretch{1.5}	
      \begin{tabular}{c | c c c | c c}
        \toprule[0.5mm]
        \textbf{Method}& \textbf{PFM-TFF} & \textbf{PFM-MFF-Uni} & \textbf{PFM-MFF-Mul} & \textbf{HOTA}& \textbf{MOTA}\\
        \hline
        \textrm{I}  & &  & &39.045 &30.585\\
        \textrm{II}  &\Checkmark &  & &40.301  &39.52   \\
        \textrm{III}  &&\Checkmark  & &  39.292 &31.497\\
        \textrm{IV} &  &  &\Checkmark &39.901 &32.783 \\ 
        \textrm{V}& &\Checkmark &\Checkmark &39.300 &35.067\\        
        \textrm{VI}  &\Checkmark &\Checkmark &\Checkmark  &41.068 &43.088\\   
        \hline
  \end{tabular}}
\label{tb::AS}

\end{table}
\subsubsection{Ablation study}
To verify the effectiveness of each component of the proposed method, we conduct some ablation experiments. The \textbf{PFM-TFF} indicates that our tracker only uses the temporal feature fusion sub-module of progressive fusion module (the first fusion stage of PFM). The \textbf{PFM-MFF-Uni} represents using fused features (as K,V) to enhance the features of two single modalities (as Q) in the second fusion stage of PFM. Similarly, the \textbf{PFM-MFF-Mul} represents using the features of two single modalities (as K,V)  to enhance fused features (as Q) in the second fusion stage of PFM. 

As shown in Table~\ref{tb::AS}, compared with baseline (\textbf{Method I}), \textbf{Method II}, \textbf{Method III} and \textbf{Method IV} have significant performance improvements on HOTA/MOTA metrics, which demonstrate that the sub-module of two fusion stages of the PFM are valid. Compared with \textbf{Method III} and \textbf{Method IV}, the \textbf{Method V} attempts to enhance both modality-specific features and fused features simultaneously. It can be seen that combining the two types of interactions has a significant tracking performance improvement on MOTA metric. In addition, it can be concluded from \textbf{Method VI} that combining all modules achieves a greater performance improvement.
\begin{table}[htbp]
  \footnotesize
  \renewcommand\arraystretch{1.8}
  \caption{Ablation study of temporal feature fusion sub-module.}
    \setlength{\tabcolsep}{1.3mm}{\begin{tabular}{c|c|lll}
    \hline
    \multicolumn{1}{p{4.315em}|}{\textbf{$+$HeatMap}} & \multicolumn{1}{p{4.19em}|}{\textbf{Methods}}& \multicolumn{1}{p{4.19em}}{\textbf{HOTA}} & \multicolumn{1}{p{4.19em}}{\textbf{IDF1}} & \multicolumn{1}{p{4.19em}}{\textbf{MOTA}} \\
    \hline
    {\checkmark} & Baseline &  39.045   & 44.42      &30.585  \\ \hline   
    {\checkmark} &  PFM-TFF & 40.301   & 46.494      &39.52  \\
  $\times$ &  PFM-TFF &  37.71 \textcolor{red}{\textbf{(-2.591)}}   &  40.996\textcolor{red}{\textbf{(-5.498)}}     & 41.388 \textcolor{green}{\textbf {(+1.868)}} \\
   \hline
    \end{tabular}}%
  \label{tab:addlabel}%
\end{table}%

To verify the effectiveness of heat map of temporal feature fusion sub-module of progressive fusion module, we conduct an ablation studies. Specifically, we show the results of our experimental analysis for the heat map in the temporal fusion module in Table~\ref{tab:addlabel}. From these results, it can be seen that the performance of our method decreases significantly after removing the heat Map, especially most noticeably in the IDF1 metric. This phenomenon suggests that the heat Map can indeed provide the tracker with valid a prior information about the target location, which is crucial for significantly improving the tracking performance.

\subsection{ Qualitative Evaluation}
\begin{figure}[t]
\includegraphics[width=0.47\textwidth]{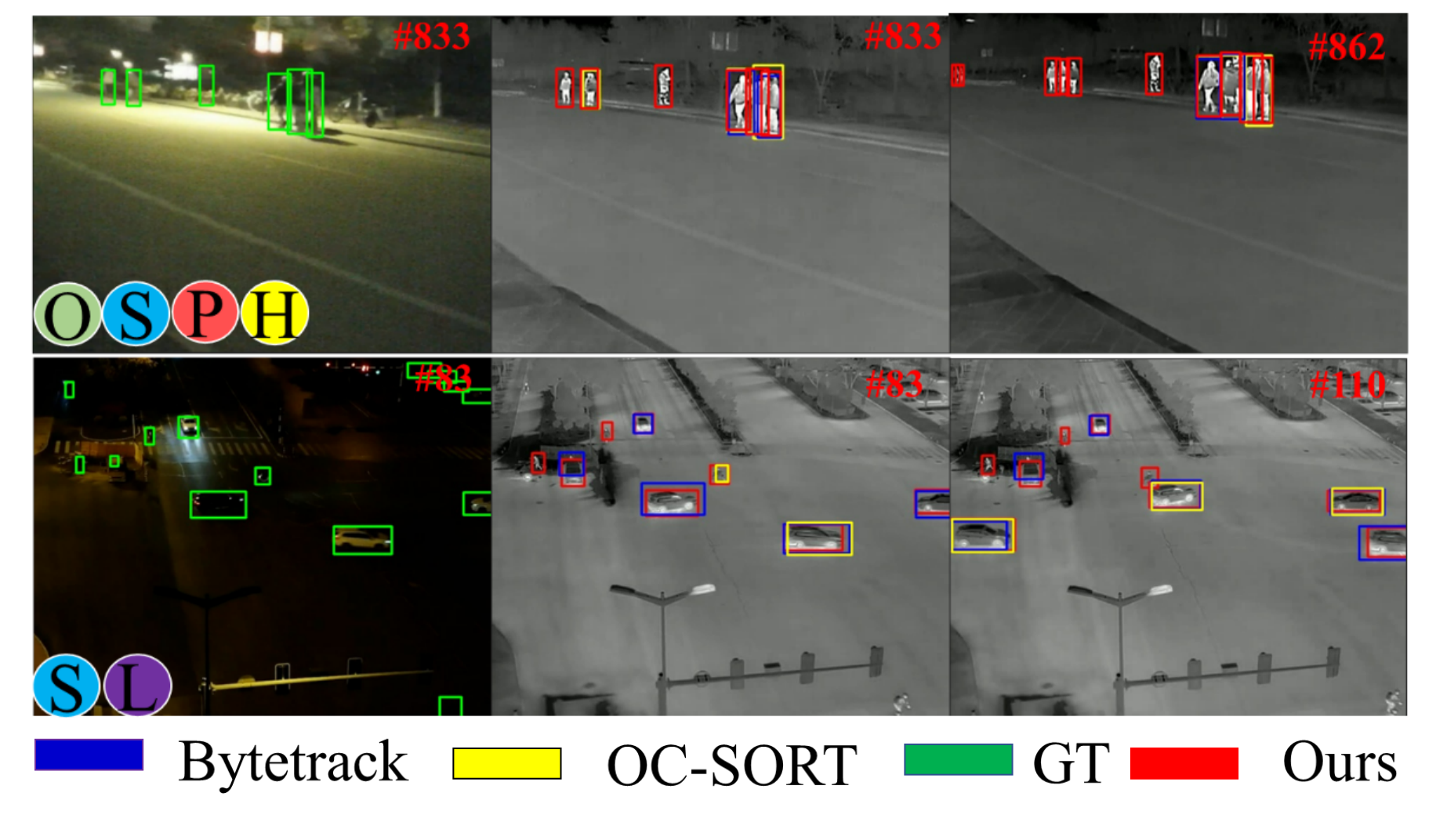}
   \caption{Qualitative comparison in VT-MOT testing set.
Challenge attributes are shown at bottom, including similar Object (O), Low illumination (L), High illumination (H), Partial occlusion (P) and Small object (S).}
   \label{fig:quantitative}

\end{figure}

As shown in Fig. ~\ref{fig:quantitative}, we compare the tracking results of our tracker with two other trackers in two sequential partial frames selected from the test set sequences "UVA-0305-17" and "photo-0318-46". 
In the lower half of Fig.~\ref{fig:quantitative}, there are challenges such as low illumination and small objects in the video sequence. In this case, ByteTrack and OC-SORT suffer from severe leakage. However, our method is able to deal with these challenges effectively. Its core advantage is that the method can effectively fuse complementary information between different modalities and can well extract contextual information of small objects, thus enhancing the capability of object tracking.
 
In the upper half of Fig.~\ref{fig:quantitative}, there are challenges such as high illumination interference, small objects, partial occlusion and similar object in the video sequence. Compared to OC-SORT and ByteTrack, our method still gives good tracking results in such a complex hybrid challenge scenario.
\section{Concluding Remarks}
In this paper, we build a large-scale visible-thermal video benchmark for MOT, which includes 582 video sequence pairs with 401k frame pairs and 3.99 million annotation boxes. The VT-MOT is collected on surveillance, UAV and handheld platforms, aiming to promote the development of MOT in multiple platforms. All video sequences are manually aligned in time and space frame by frame. Moreover, we propose a simple yet effective tracking framework, which effectively fuses temporal information and complementary information of two modalities in a progressive manner, for robust visible-thermal MOT. A large number of experiments are conducted on VTUAV and the results prove the superiority and effectiveness of the proposed method compared with state-of-the-art methods. 

In the future, there are several potential research directions in visible-thermal MOT, as follows: 

 \noindent{\bf Development of a lightweight and high performance visible-thermal MOT algorithm:} In the field of MOT, balancing algorithmic efficiency and accuracy is challenging. High-accuracy tracking methods are usually inefficient, and the introduction of multimodal data improves accuracy but further increases computational burden. Mamba network~\cite{gu2023mamba} is in the spotlight for high performance, fast training, and fast inference. Therefore, efficient utilization of Mamba structure to balance accuracy and efficiency is important for the advancement of MOT technology.
 
 \noindent{\bf Utilizing large models:} MOT encounter many limitations in algorithm performance due to the complexity of its tasks. In recent years, the emergence of large model, prompt and adapter techniques bring new hope to the development of MOT. Utilizing the capability of large models through prompt or adapter techniques is the trend of visible and thermal MOT development. Some works~\cite{zhu2023visual, hong2024onetracker} can be referred to do multimodal prompt or adapter learning.
 
 \noindent{\bf Similar objects in thermal infrared modality:} Thermal infrared lacks color and texture, making it challenging to distinguish similar objects, especially human targets with minor shape differences. To overcome this, joint modeling of appearance and trajectory position is essential, along with a robust fusion method to integrate discriminative features from both modalities.
 
 \noindent{\bf Modal unaligned tracking:} Real-world multimodal data often suffer from misalignment. One approach is to develop alignment-free algorithms using the Transformer for feature interaction. Another approach focuses on constructing networks that align, fuse, and track synergistically, addressing local region inconsistencies through global and local alignment methods. Some works~\cite{shi2022rethinking, truong2020glu, xu2022rfnet} can be referenced. 
 
 \noindent{\bf Multi-task development of VT-MOT:} The dataset can support multiple tasks such as video detection, object detection from video frames, and cross-modal or unimodal detection and tracking, which can be explored in future developments.

\bibliographystyle{IEEEtran}
\bibliography{sample-base}

\vfill

\end{document}